\documentclass[10pt,twocolumn,letterpaper]{article}

\usepackage{cvpr}
\usepackage{times}
\usepackage{epsfig}
\usepackage{graphicx}
\usepackage{amsmath}
\usepackage{amssymb}
\usepackage{subfigure}
\usepackage{url}
\usepackage{booktabs}
\usepackage{graphicx}
\usepackage{float}
\usepackage{stfloats}
\usepackage{cases}

\usepackage[pagebackref=true,breaklinks=true,letterpaper=true,colorlinks,bookmarks=false]{hyperref}

\cvprfinalcopy 

\newif\ifshowcomments
\showcommentstrue

\ifshowcomments

\newcommand{\TODO}[1]{{\color{red}{[TODO: #1]}}}

\newcommand{\revised}[1]{{\color[rgb]{0.2,0.7,0.2}{#1}}}

\newcommand{\phil}[1]{{\color[rgb]{0.8,0.1,0.1}{#1}}}

\else
\newcommand{\TODO}[1]{}
\newcommand{\revised}[1]{}
\newcommand{\phil}[1]{}
\fi

\DeclareMathOperator{\atantwo}{atan2}

\def\cvprPaperID{2413} 

\ifcvprfinal\fi
\begin{document}

\title{Instance Shadow Detection}

\author{Tianyu Wang$^{1,2, \ast}$, Xiaowei Hu$^{1,}$\thanks{Joint first authors}\ , Qiong Wang$^{2}$, Pheng-Ann Heng$^{1,2}$, and Chi-Wing Fu$^{1,2}$\\
$^1$ Department of Computer Science and Engineering, The Chinese University of Hong Kong\\
$^2$ Shenzhen Key Laboratory of Virtual Reality and Human Interaction Technology, \\ Shenzhen Institutes of Advanced Technology, Chinese Academy of Sciences, China
}

\maketitle
\thispagestyle{empty}


\begin{abstract}

Instance shadow detection is a brand new problem, aiming to find shadow instances paired with object instances.
%
%
%
To approach it, we first prepare a new dataset called SOBA, named after Shadow-OBject Association, with 3,623 pairs of shadow and object instances in 1,000 photos, each with individually-labeled masks.
Second, we design LISA, named after Light-guided Instance Shadow-object Association, an end-to-end framework to automatically predict the shadow and object instances, together with the shadow-object associations and light direction.
Then, we pair up the predicted shadow and object instances and match them with the predicted shadow-object associations to generate the final results.
In our evaluations, we formulate a new metric named the shadow-object average precision to measure the performance of our results.
Further, we conducted various experiments and demonstrate our method's applicability to light direction estimation and photo editing.




%
%
%
%

\end{abstract}



\section{Introduction}


\noindent
``{\em When you light a candle, you also cast a shadow,\/}''---Ursula K. Le Guin written in A Wizard of Earthsea.

When some objects block the light, shadows are formed.
And when we see a shadow, we also know that there must be some objects that create or cast the shadow.
Shadows are light-deficient regions in a scene, due to light occlusion, but they carry the shape of the light-occluding objects, as they are projections of these objects onto the physical world.
In this work, we are interested in a new problem,~\ie, {\em finding shadows together with their associated objects\/}.


Concerning shadows, prior works in computer vision and image understanding focus mainly on shadow detection~\cite{hou2019large,hu2019revisiting,Hu_2018_CVPR,khan2014automatic,khan2016automatic,le2018a+d,vicente2016large,zheng2019distraction,zhu2018bidirectional} and shadow removal~\cite{ding2019argan,hu2019direction,hu2019mask,Le_2019_ICCV,qu2017deshadownet,wang2018stacked}.
Our goal in this work is to leverage the remarkable computation capability of deep neural networks to address the new problem of associating shadows and objects---{\em instance shadow detection\/}.
That is, we want to detect the shadow instances in images, together with the associated object that casts each shadow.

Being able to find shadow-object associations has the potentials to benefit various applications.
For example, for privacy protection, when we remove humans and cars from photos, we can remove objects and associated shadows altogether.
In a recent work on removing objects from images for privacy protection~\cite{Uittenbogaard2019privacy}, the shadows are simply left behind.
Also, when we edit photos, say by scaling or translating objects, we can naturally manipulate objects with their associated shadows simultaneously.
Further, shadow-object associations give hints to the light direction in the scene, supporting applications such as relighting.

\begin{figure}[!t]
\centering
\includegraphics[width=0.99\linewidth]{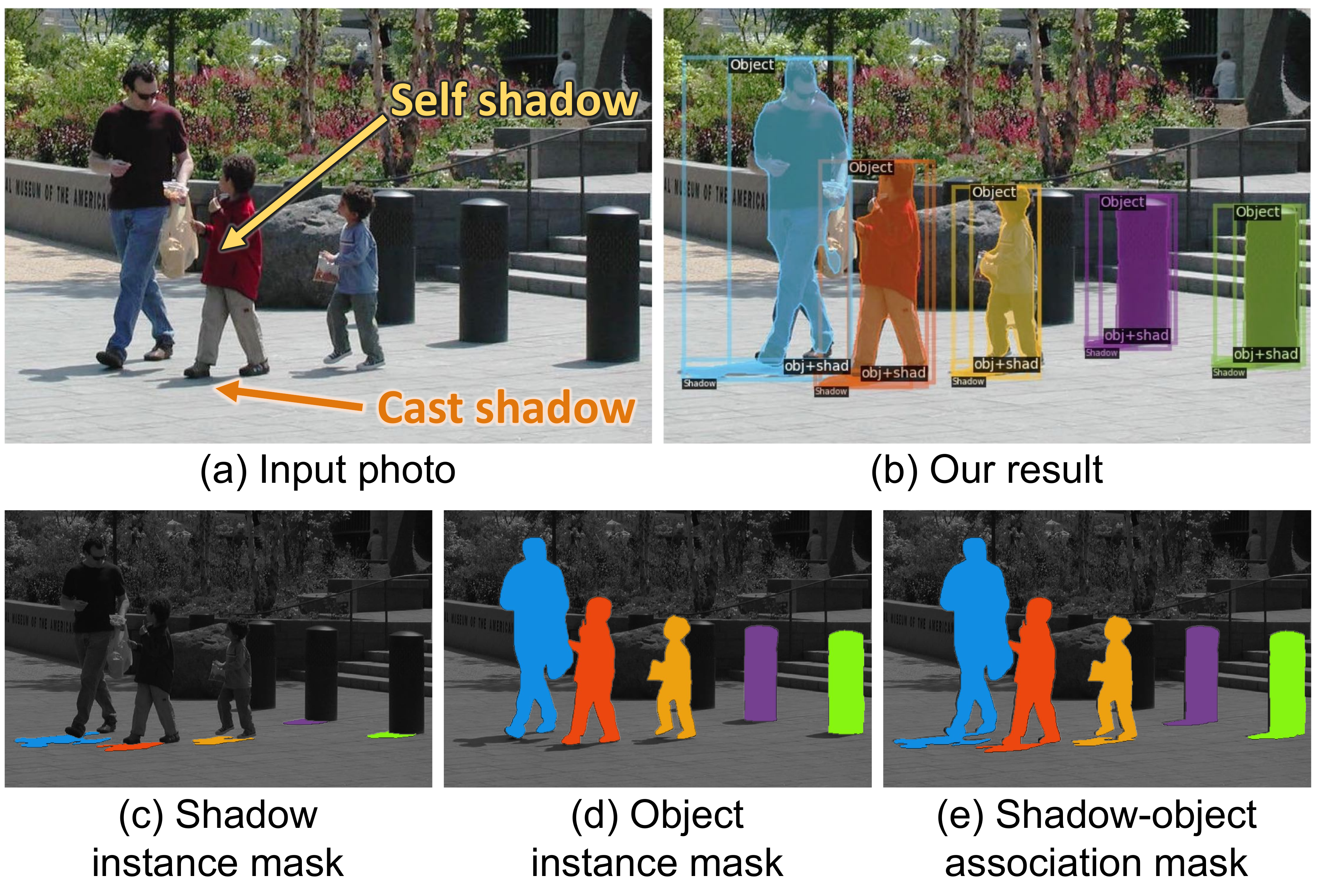}
\caption{Given a photo with shadows (a), the problem of {\em instance shadow detection\/} is to detect the individual shadow instances (c) and the individual object instances (d), as well as to associate the shadows with the objects (e) that cast them.
(b) shows the prediction results produced by our method on (a).}
\label{img:1}
\vspace*{-3mm}
\end{figure}

\begin{figure*}[t]
\centering
\includegraphics[width=0.99\textwidth]{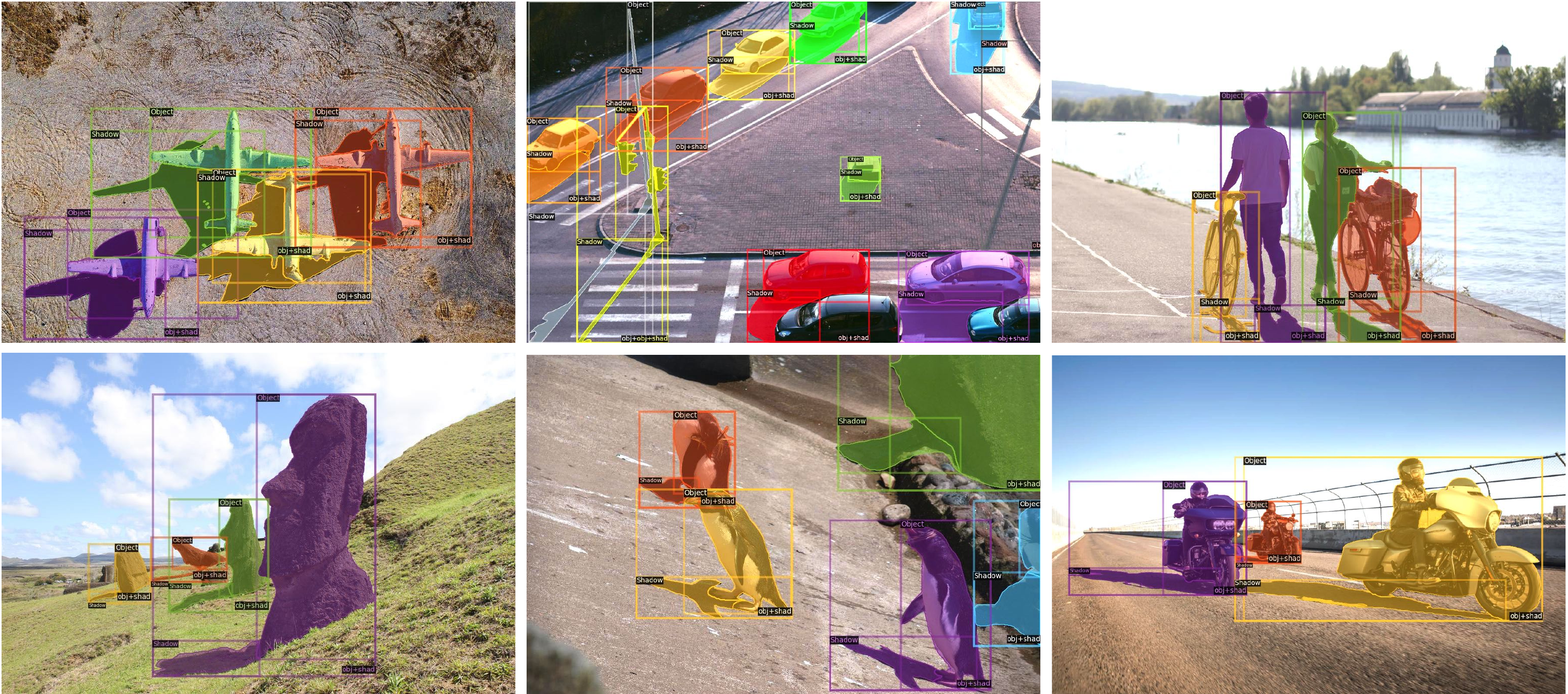}
\caption{Example images with mask and box labels in our SOBA dataset.
Please zoom in for better visualization.}
\label{img:dataset}
\vspace*{-1.5mm}
\end{figure*}

%
To approach the problem of instance shadow detection, 
first, we prepare a new dataset called \emph{SOBA}, named after \emph{Shadow OBject Association}.
SOBA contains 3,623 pairs of shadow-object associations over 1,000 photos, each with three masks (see Figures~\ref{img:1} (c)-(e)):
(i) shadow instance mask, where we label each shadow instance with a unique color; 
(ii) shadow-object association mask, where we label each shadow-object pair with a corresponding unique color; and
(iii) object instance mask, which is (ii) minus (i).
%
In general, there are two types of shadows:
(i) {\em cast shadows\/}, formed on background objects, usually ground, as the projections of the light-occluding objects, and
(ii) {\em self shadows\/}, formed on the side of the light-occluding objects opposite to the direct light (see Figure~\ref{img:1}(a)).
%
In this work, we consider mainly cast shadows, which are object projections, since self shadows are already on the associated objects.
See also Figure~\ref{img:dataset} for example images in our SOBA dataset.

Next, we design an end-to-end framework called \emph{LISA}, named after {\em Light-guided Instance Shadow-object Association\/}, to find 
the individual shadow and object instances, the shadow-object associations, and the light direction in each shadow-object association.
%
From these predictions, we then use a simple yet effective method to pair the predicted shadow and object instances and to match them with the predicted shadow-object associations.

Third, to quantitatively measure and evaluate the performance of the instance shadow detection results, we formulate a new evaluation metric called SOAP, named after {\em Shadow-Object Average Precision\/}.
In the end, we further perform a series of experiments to show the effectiveness of our method and demonstrate its applicability to light direction estimation and photo editing.

\section{Related Work}

\paragraph{Shadow detection.} \
%
Early works~\cite{salvador2004cast,panagopoulos2011illumination,tian2016new} made use of physical illumination and color models, and analyzed the spectral and geometrical properties of shadows.
Later, machine learning methods were explored to detect shadows by modeling shadows based on handcrafted features,~\eg, texture~\cite{zhu2010learning,vicente2015leave,guo2011single,vicente2018leave}, color~\cite{lalonde2010detecting,vicente2015leave,guo2011single,vicente2018leave}, T-junction~\cite{lalonde2010detecting}, and edge~\cite{lalonde2010detecting,zhu2010learning,huang2011characterizes}, then by using various classifiers,~\eg, decision tree~\cite{lalonde2010detecting,zhu2010learning} and SVM~\cite{guo2011single,huang2011characterizes,vicente2015leave,vicente2018leave}, to differentiate shadows and non-shadows.
However, physical models and handcrafted features have limited feature representation capability, thus they are not robust in general situations.

Later, convolutional neural networks (CNN) were introduced to detect shadows.
Khan~\etal~\cite{khan2014automatic} and Shen~\etal~\cite{shen2015shadow} used CNN to learn high-level features and optimization methods to detect shadows.
Vicente~\etal~\cite{vicente2016large} trained a fully-connected network to predict a shadow probability map, then locally refine the shadows via a patch-CNN.

More recently, end-to-end networks were designed to detect shadows.
%
Nguyen~\etal~\cite{nguyen2017shadow} built a conditional generative adversarial network with a sensitive parameter to stabilize the network training.
%
Hu~\etal~\cite{hu2019direction,Hu_2018_CVPR} and Zhu~\etal~\cite{zhu2018bidirectional} explored the spatial context via the direction-aware spatial context module and recurrent attention residual module, respectively.
Wang~\etal~\cite{wang2018stacked} and Ding~\etal~\cite{ding2019argan} jointly detected and removed shadows by using multiple networks or a multi-branch network.
To improve the detection performance, Le~\etal~\cite{le2018a+d} proposed to generate more training samples, while Zheng~\etal~\cite{zheng2019distraction} combined the strengths of multiple methods to explicitly revise the results.
%
%
This work explores a new problem on detecting shadows, namely {\em instance shadow detection\/}.
Unlike general shadow detection, which finds only a single mask for all shadows in an image, we design a deep architecture to find not just the individual shadows but also the associated objects altogether.


\vspace*{-3.5mm}
\paragraph{Instance segmentation.} \
Besides, this work relates to the emerging problem of instance segmentation, where the goal is to label pixels of individual foreground objects in the input image.
Overall, there are two major approaches to the problem: proposal-based and proposal-free approaches.

Proposal-based approach generally uses object detectors to propose candidates and classifies the candidates to find object instances,~\eg,
MNC~\cite{dai2016instance}, DeepMask~\cite{pinheiro2015learning}, InstanceFCN~\cite{dai2016instance}, and SharpMask~\cite{pinheiro2016learning}.
%
%
%
%
Later, FCIS~\cite{li2017fully} jointly detects and segments the object instances using a fully convolutional network.
BAIS~\cite{hayder2017boundary} models the object shapes and segments the object instances in a boundary-aware manner.
MaskLab~\cite{chen2018masklab} uses a network with three outputs for box detection, semantic segmentation, and direction prediction, while 
methods based on Mask R-CNN~\cite{he2017mask},~\eg,~\cite{liu2018path,chen2019hybrid,peng2018megdet}, achieved great performance by simultaneously detecting the object instances and predicting the segmentation masks.

The proposal-free approach~\cite{arnab2017pixelwise,bai2017deep,kirillov2017instancecut,liu2017sgn} first classifies the image pixels, then group the pixels into object instances.
%
%
Recently, TensorMask~\cite{chen2019tensormask} leverages a fully convolutional network 
for dense mask prediction, while
SSAP~\cite{gao2019ssap} predicts the object instance labels in just a single pass.
\section{SOBA (Shadow OBject Association) Dataset}

\begin{figure}[tp]     
\centering
\includegraphics[width=0.495\linewidth]{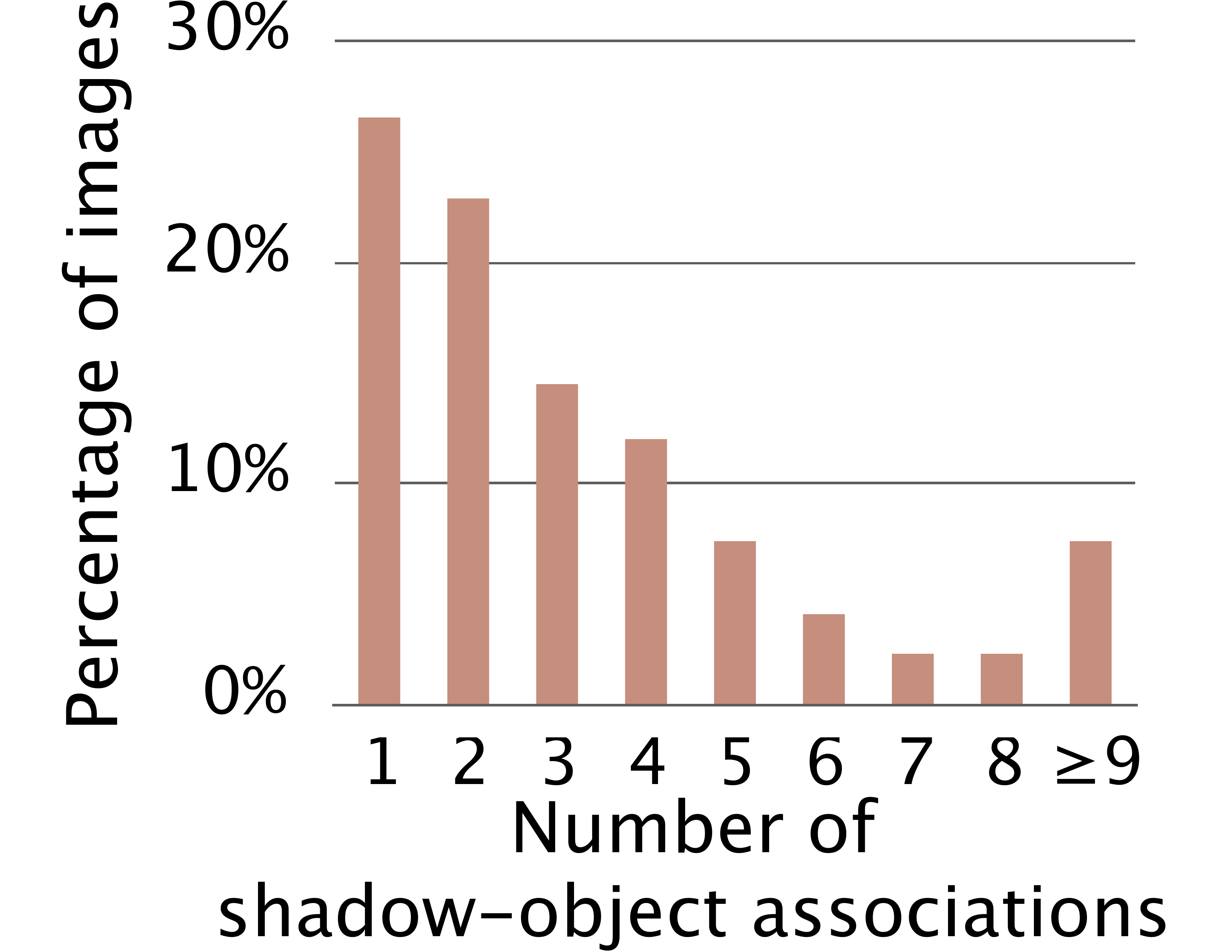}
\includegraphics[width=0.495\linewidth]{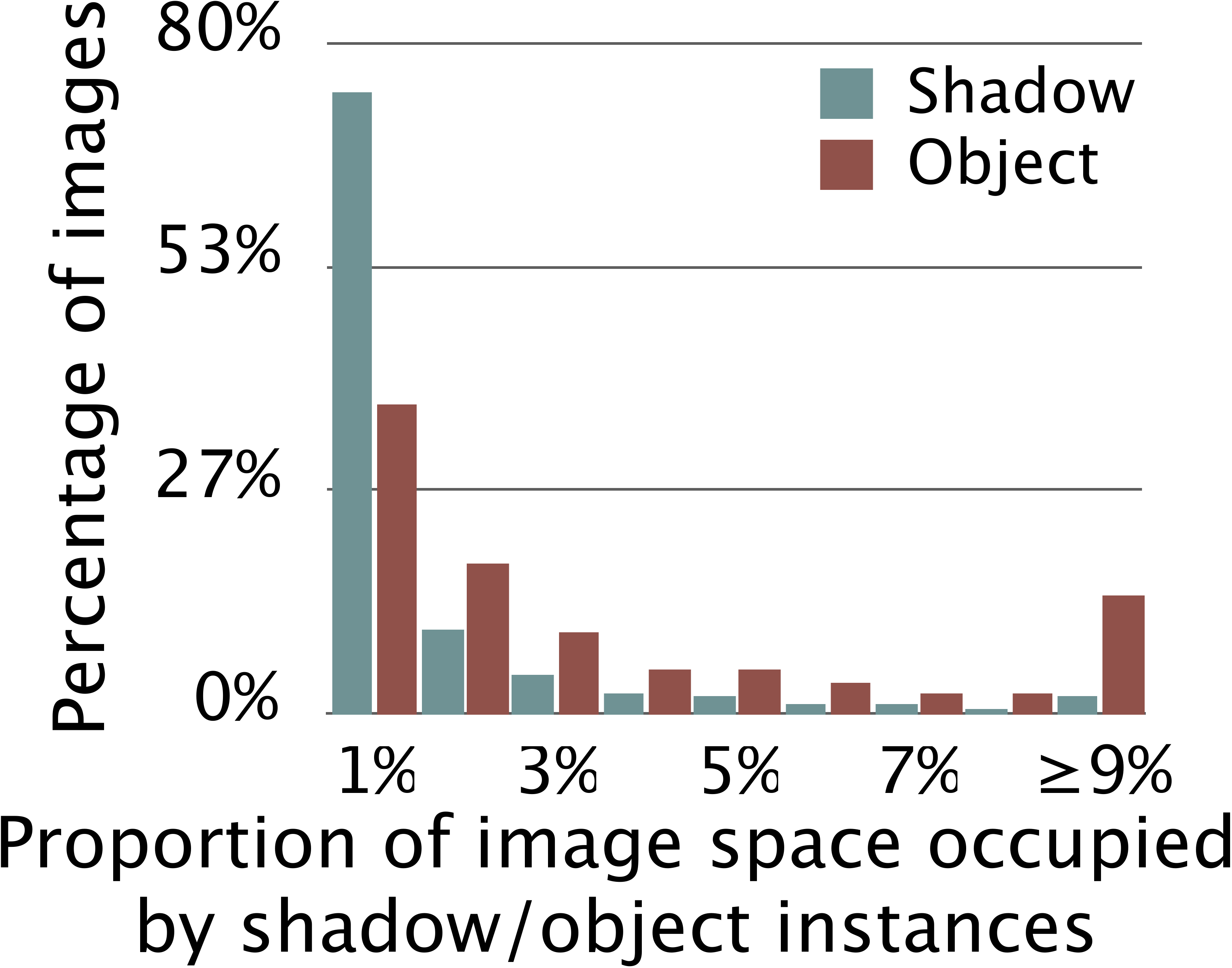}
\caption{Statistical properties of the SOBA dataset.}
\label{img:dataset_analysis}
\vspace*{-3mm}
\end{figure}

We collected 1,000 images from the ADE20K~\cite{zhou2017scene,zhou2019semantic}, SBU~\cite{hou2019large, vicente2016noisy, vicente2016large}, ISTD~\cite{wang2018stacked}, and Microsoft COCO~\cite{lin2014microsoft} datasets, and also from the Internet using keyword search with shadow plus animal, people, car, athletic meeting, zoo, street, etc.
%
Then, we coarsely label the images to produce the shadow instance masks and shadow-object association masks, and refine them using 
Apple Pencil; see Figures~\ref{img:1} (c) \& (e).
Next, we obtain the object instance masks (see Figure~\ref{img:1} (d)) by subtracting each shadow instance mask from the associated shadow-object association mask.
Overall, there are 3,623 pairs of shadow-object instances in the dataset images, and we randomly split the images into a training set (840 images, 2,999 pairs) and a testing set (160 images, 624 pairs); see Figure~\ref{img:dataset} for some examples.

\begin{figure}[!t]
	\centering
	\includegraphics[width = 0.95\linewidth]{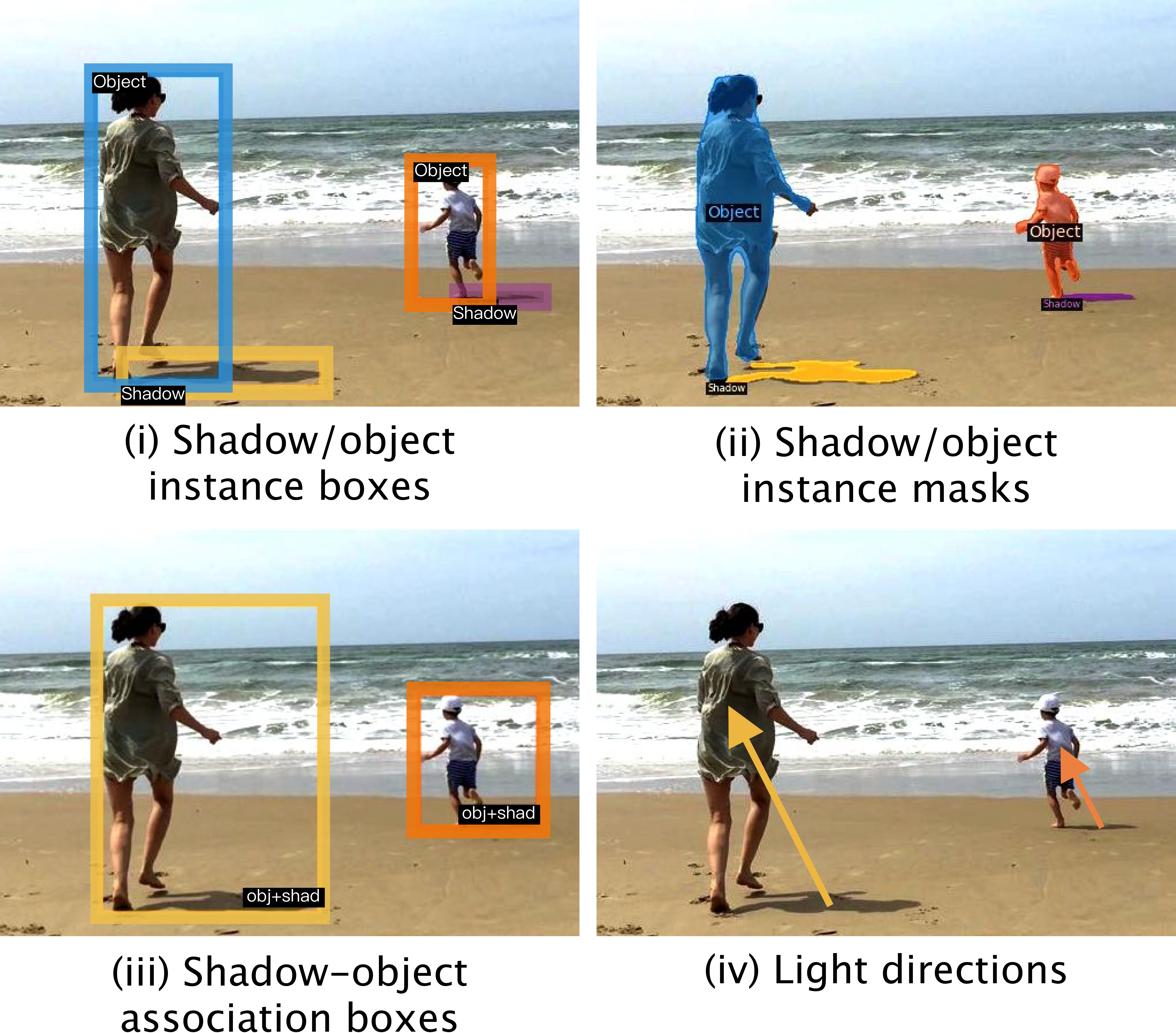} 
	\caption{Example predictions (output) from our LISA framework.}
	\label{img:output_example}
	\vspace*{-1.5mm}
\end{figure}

%
Figure~\ref{img:dataset_analysis} shows some statistical properties of the SOBA dataset.
From the histogram shown on the left, we can see that SOBA has a diverse number of shadow-object pairs per image, with around 3.62 pairs per image on average.
%
%
Also, it contains many challenging cases: 7\% of the images have nine or more shadow-object pairs per image.
%
On the other hand, the histogram shown on the right reveals the proportion of image space (horizontal axis) occupied, respectively, by the shadow and object instances in the dataset images.
From the plot, we can see that most shadows and objects occupy relatively small areas in the whole images, demonstrating the challenges to detect them.


%
%

%


\section{Methodology}


\subsection{Overall Network Architecture of LISA}

\begin{figure*}[tp]     
\centering
\includegraphics[width = 0.95\linewidth]{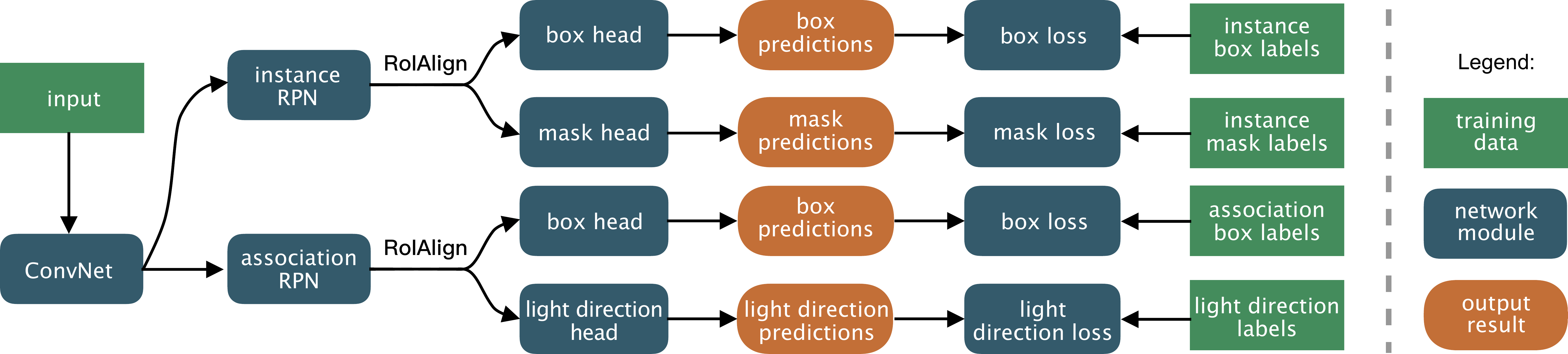}
\caption{The schematic illustration of our Light-guided Instance Shadow-object Association (LISA) framework.}
\label{img:network}
\vspace*{-1.5mm}
\end{figure*}

Compared with shadow detection, the challenges of instance shadow detection are that we have to predict shadow instances rather than just a single mask for all the shadows in the input image.
Also, we have to find object instances in the input image and pair them up with the shadow instances.
To meet these challenges, we design an end-to-end framework called LISA, named after Light-guided Instance Shadow-object Association.
Overall, as shown in Figure~\ref{img:network}, LISA takes a single image as input and predicts
\begin{itemize}
\vspace*{-1mm}
\item[(i)]
a box of each shadow/object instance,
\vspace*{-1.5mm}
\item[(ii)]
a mask of each shadow/object instance,
\vspace*{-1.5mm}
\item[(iii)]
a box of each shadow-object association (pair), and
\vspace*{-1.5mm}
\item[(iv)]
the light direction for each shadow-object association.
\end{itemize}
Figure~\ref{img:output_example} shows a set of example outputs.
Particularly, LISA predicts the light direction and takes it as guidance to find shadow-object associations, since the light direction is usually consistent with the shadow-object associations.
%



Figure~\ref{img:network} shows the architecture of LISA, which begins by using a convolutional neural network (ConvNet) to extract semantic features from the input image.
Here, we use the feature pyramid network~\cite{Lin2017fpn} as the backbone ConvNet.
%
Then, we design a two-branch architecture: the top branch predicts the box and mask for each shadow/object instance and the bottom branch predicts the box for each shadow-object association and the associated light direction.

In detail, the top branch starts with the instance region proposal network (RPN)~\cite{ren2015faster} to find region proposals, which are regions with high probabilities of containing the shadow/object instances.
Then, we adopt RoIAlign~\cite{he2017mask} to extract features for each proposal and leverage the box and mask heads to predict the boxes and masks for the shadow and object instances by minimizing the loss between the prediction results and the supervision signals from the training data.
Please refer to Mask R-CNN~\cite{he2017mask} for the detail.
%
On the other hand, the bottom branch adopts an association RPN to generate region proposals for the shadow-object associations, then uses RoIAlign to extract features for each proposal and adopts the box head to produce the bounding boxes of the shadow-object associations.
After obtaining the associations, we can then efficiently obtain the masks of the shadow-object associations by combining the shadow and object masks predicted from the top branch.
Note that the parameters in the box head are learned by minimizing the loss between the boxes of the predicted shadow-object associations and the ground-truth associations.

Besides, we design a light direction head in parallel with the box head of the bottom branch to predict an angle that represents the estimated light direction from shadow to object in each association pair.
Note that we compute the ground-truth angle $\theta^g$ of the light direction by
\begin{equation}
\nonumber
\theta^g \ = \ \atantwo( \ y_o^g - y_s^g  , \  x_o^g - x_s^g \ ) \ , 
\end{equation} 
where $(x_s^g, y_s^g)$ and $(x_o^g, y_o^g)$ are 2D coordinates of the shadow and object instance centroids in the ground-truth image, and $\atantwo(y,x)$ is a variation of the $\arctan(y/x)$ function to avoid anomaly and output a full-range polar angle in $(-\pi,\pi]$.
The shadow-object association branch and light direction branch share the common feature extraction network and the association RPN. 
By jointly optimizing the predictions of the light direction and shadow-object association in each region proposal, we can improve the overall performance of instance shadow detection; see the experimental results in Section~\ref{sec:experiments}.

\subsection{Pairing up Shadow and Object Instances}
\label{sec:4.2}

The raw predictions of LISA include shadow instances, object instances, shadow-object associations, and a light direction predicted per association.
Note that, the predicted shadow and object instances are not paired, whereas the predicted shadow-object associations are not separated as shadows and objects.
Also, some of these predictions may not be correct, and they may also contradict one another.
Hence, we have to analyze these predictions, pair up the predicted shadow and object instances, and match them with the predicted shadow-object associations, so that we can find and output the final paired shadow and object instances.

\begin{figure*}[!t]
	\centering
	\includegraphics[width=0.97\textwidth]{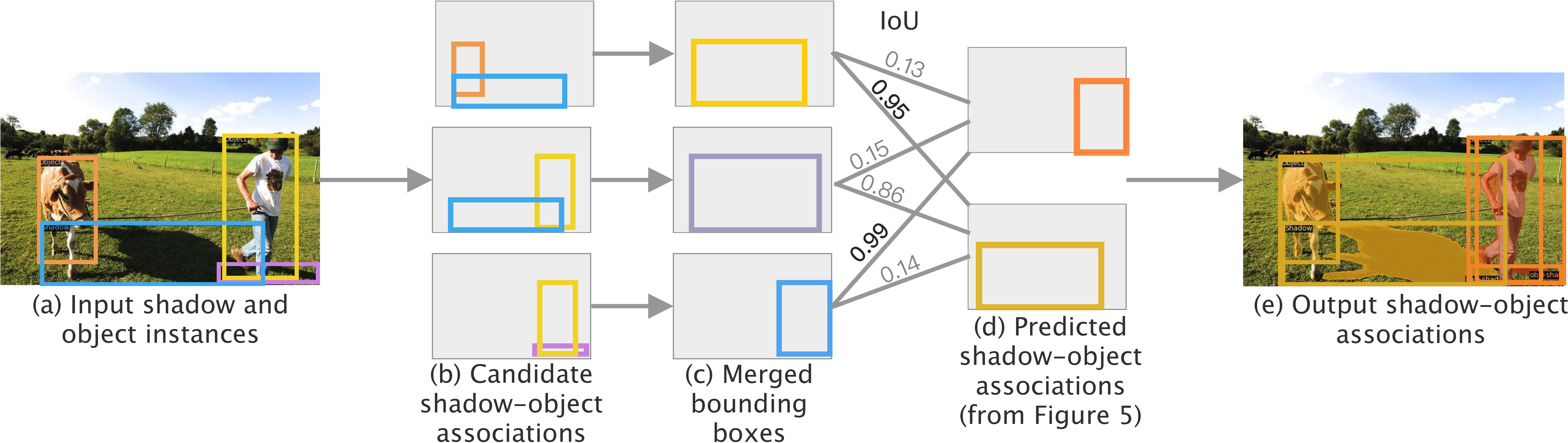}
	\caption{The pair-and-match procedure for pairing the predicted shadow and object instances and efficiently matching them with the predicted shadow-object associations.}
	\label{img:matchor}
	\vspace*{-3mm}
\end{figure*}

Figure~\ref{img:matchor} illustrates the procedure, where we first find candidate shadow-object associations (see Figure~\ref{img:matchor} (b)) by
(i) computing the shortest distance between the bounding boxes of every pair of shadow and object instances, and 
(ii) regarding a pair as a candidate association, if the computed distance is smaller than a threshold, which is empirically set as the height of the associated shadow instance.
After that, we construct bounding box $B_i$ for the $i$-th candidate pair (see Figure~\ref{img:matchor} (c)) by merging the bounding boxes of the associated shadow and object instances.
Given ($x^s_{\min}$,$y^s_{\min}$) and ($x^s_{\max}$,$y^s_{\max}$) as the lower-left and upper-right corners of the shadow instance bounding box, and ($x^o_{\min}$,$y^o_{\min}$) and ($x^o_{\max}$,$y^o_{\max}$) as the lower-left and upper-right corners of the object instance bounding box, the corners of the merged bounding box $B_i$ are given by
\begin{eqnarray}
\begin{array}{c@{\hspace*{1mm}}c@{\hspace*{1mm}}c@{\hspace*{1mm}}c@{\hspace*{1mm}}c@{\hspace*{1mm}}c}
& \big(
& \min( x^s_{\min} , x^o_{\min} ) & ,
& \min( y^s_{\min} , y^o_{\min} ) & \big) \ ,
\nonumber
\\
\text{and} \
& \big(
& \max( x^s_{\max} , x^o_{\max} ) & ,
& \max( y^s_{\max} , y^o_{\max} ) & \big) \ .
\end{array}
\end{eqnarray}

In the end, as illustrated in Figure~\ref{img:matchor} (d), we compute the Intersection over Union (IoU) between the merged boxes and the shadow-object association boxes predicted independently in LISA (see Figure~\ref{img:network}), and select those with the highest IoUs as the final shadow-object associations.
Then, for each of these associations, we can get back the associated shadow instance and object instance, and pair them as the final outputs; see Figure~\ref{img:matchor} (e).

\subsection{Training Strategies}

\vspace*{1mm}
\paragraph{Loss function.}
We optimize LISA by jointly minimizing the instance box loss, instance mask loss, association box loss, light direction loss (see Figure~\ref{img:network}), and the losses of instance RPN and association RPN.
The loss functions of boxes, masks, and RPNs follow the formulations in Mask R-CNN~\cite{he2017mask}, whereas the light direction loss $L_{light}$ is formulated by a smooth $L_1$ loss~\cite{girshick2015fast}, as follows:
\begin{equation}
\nonumber
L_{light}(\theta^p, \theta^g) \ = \
\left\{
\begin{array}{ll}
0.5 \ (\theta^p - \theta^g)^2  & \text{if} \  |\theta^p - \theta^g|<1   \\
|\theta^p - \theta^g| - 0.5    & \text{otherwise},   
\end{array}
\right.
\end{equation}
where $\theta^p$ and $\theta^g$ are the predicted and ground-truth angles of the light direction, respectively. 

\vspace*{-3mm}
\paragraph{Training parameters.}
We train our LISA framework by following the training strategies of Mask R-CNN implemented on Facebook Detectron2~\cite{wu2019detectron2}.
Specifically, we adopt the weights of ResNeXt-101-FPN~\cite{Lin2017fpn,xie2017aggregated} trained on ImageNet~\cite{deng2009imagenet} to initialize the parameters of the backbone network, and train our framework on two GeForce GTX 1080 Ti GPUs (four images per GPU) for 40$k$ training iterations. 
We set the base learning rate as 1e-4, adopt a warm-up~\cite{goyal2017accurate} strategy to linearly increase the learning rate to 1e-3 during the first 1,000 iterations, keep the learning rate as 1e-3, and stop the learning after 40$k$ iterations.
We re-scale the input images, such that the longer side is less than 1,333 and the shorter side is less than 800 without changing the image aspect ratio.
Lastly, we randomly apply horizontal flips on the images for data augmentation.

\begin{figure*}[tp]
	\centering
	\includegraphics[width=0.99\linewidth]{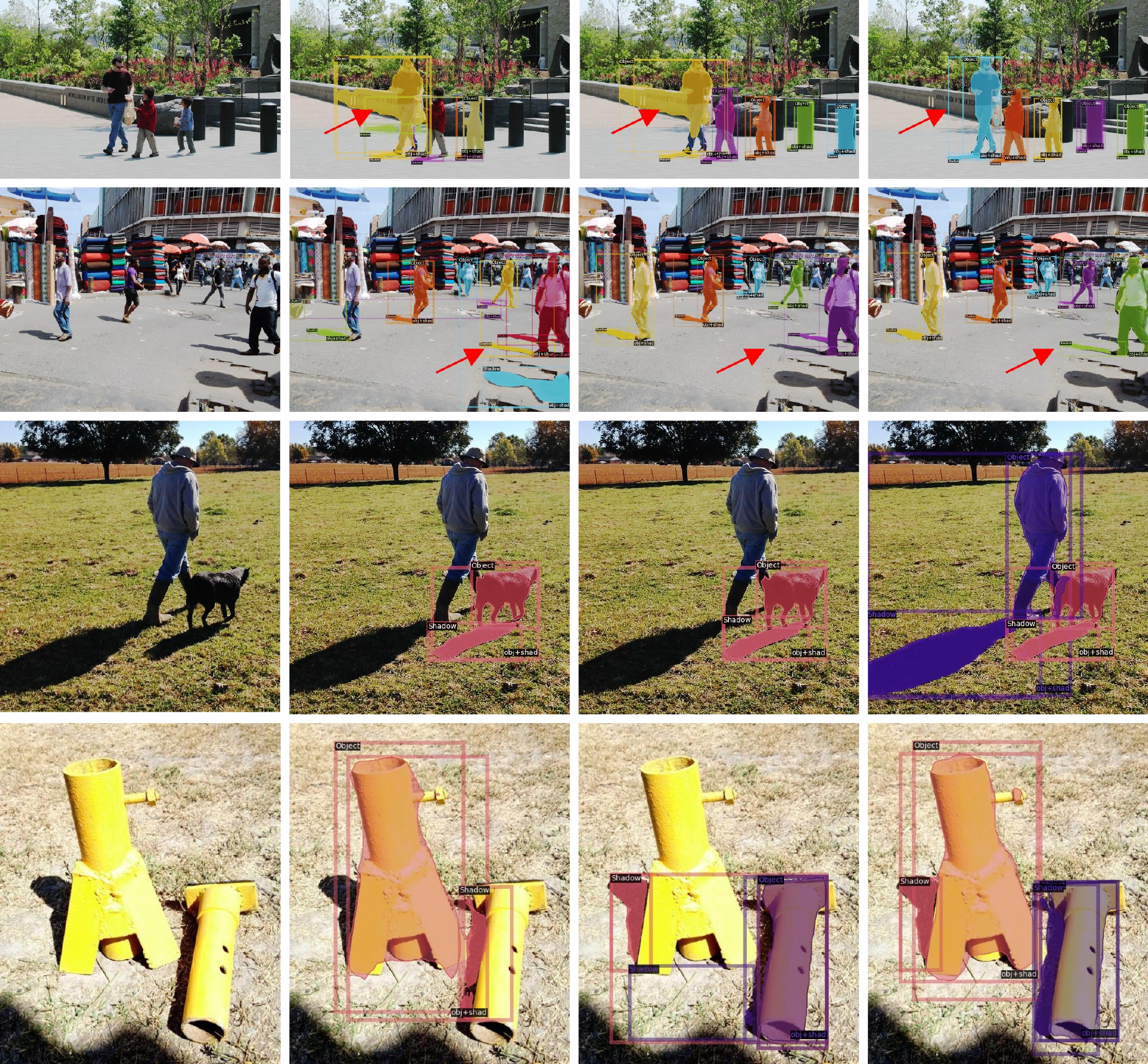}
	\begin{minipage}[t]{0.23 \linewidth}
		\vspace*{-3mm}
		\centerline{\footnotesize (a) Input image}
	\end{minipage}
	\begin{minipage}[t]{0.25 \linewidth}
		\vspace*{-3mm}
		\centerline{\footnotesize (b) Baseline 1}
	\end{minipage}
	\begin{minipage}[t]{0.26 \linewidth}
		\vspace*{-3mm}
		\centerline{\footnotesize (c) Baseline 2}
	\end{minipage}
	\begin{minipage}[t]{0.24 \linewidth}
		\vspace*{-3mm}
		\centerline{\footnotesize (d) Full pipeline}
	\end{minipage}
	\caption{Visual comparison of instance shadow detection results produced by our full pipeline and two other baseline frameworks.}
	\label{img:res1}
	\vspace*{-1.5mm}
\end{figure*}

\section{Experiments}
\label{sec:experiments}



\subsection{Evaluation Metrics}

Existing metrics evaluate instance segmentation results by looking at object instances individually.
Our problem involves multiple types of instances: shadows, objects, and their associations.
Hence, we formulate a new metric called the \emph{Shadow-Object Average Precision} (SOAP) by adopting the same formulation as the traditional average precision (AP) with the intersection over union (IoU) but further considering a sample as true positive (an output shadow-object association), if it satisfies the following three conditions:
\begin{itemize}
\vspace{-1mm}
\item[(i)]
the IoU between the predicted shadow instance and ground-truth shadow instance is no less than $\tau$;
\vspace{-1.5mm}
\item[(ii)]
the IoU between the predicted object instance and ground-truth object instance is no less than $\tau$; and
\vspace{-1.5mm}
\item[(iii)]
the IoU between the predicted and ground-truth shadow-object associations is no less than $\tau$. 
\end{itemize}

We follow~\cite{lin2014microsoft} to report the evaluation results by setting $\tau$ as 0.5 ($\text{SOAP}_{50}$) or 0.75 ($\text{SOAP}_{75}$), and also report the average over multiple $\tau$ [0.5:0.05:0.95] ($\text{SOAP}$).
Moreover, since we can obtain the bounding boxes as well as the masks for the shadow instances, object instances, and shadow-object associations, we further report $\text{SOAP}_{50}$, $\text{SOAP}_{75}$, and $\text{SOAP}$ in terms of both bounding boxes and masks.

%




\begin{table}[tp]
\centering
\caption{Comparing our full pipeline with two simplified baseline frameworks on the bounding boxes of the final shadow-object associations in terms of SOAP$_{50}$, SOAP$_{75}$, and SOAP.}
\vspace{1mm}
\label{tab:SOAP_box}
\resizebox{0.92\linewidth}{!}{%
\begin{tabular}{cccc}
\toprule
Method & box SOAP$_{50}$ & box SOAP$_{75}$ & box SOAP \\ \midrule
Baseline 1 & 40.3 & 14.0 & 16.7 \\
Baseline 2 & 47.8 & 14.0 & 19.6 \\
Our full pipeline &\textbf{50.5} & \textbf{16.4} & \textbf{21.8} \\ \midrule
\end{tabular}%
}
\vspace*{-1mm}
\end{table}

\begin{table}[tp]
\centering
\caption{Comparing our full pipeline with two simplified baseline frameworks on the masks of the final shadow-object associations in terms of SOAP$_{50}$, SOAP$_{75}$, and SOAP.}
\vspace{1mm}
\label{tab:SOAP_mask}
\resizebox{0.98\linewidth}{!}{%
\begin{tabular}{cccc}
\toprule
Method & mask SOAP$_{50}$ & mask SOAP$_{75}$ & mask SOAP \\ \midrule
Baseline 1 & 41.0 & 10.0 & 16.7 \\
Baseline 2 & 48.1 & 12.5 & 20.1 \\
Our full pipeline & \textbf{50.9} & \textbf{14.4} & \textbf{21.6} \\ \bottomrule
\end{tabular}%
}
\vspace*{-2mm}
\end{table}

\begin{figure*}[tp]
	\centering
	\includegraphics[width=0.999\linewidth]{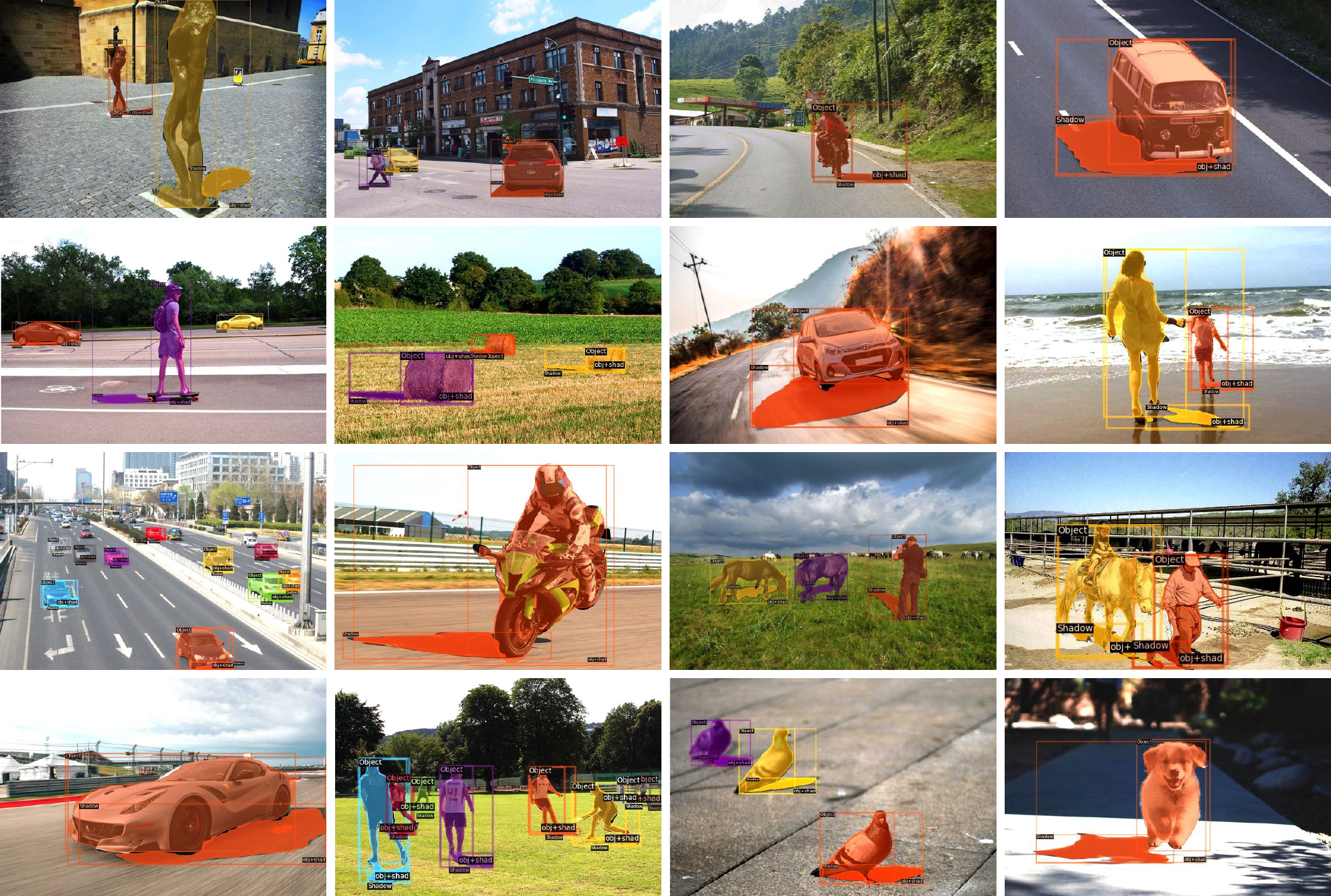}
	\caption{Instance shadow detection results produced by our method over a wide variety of photos and objects.}
	\label{img:more_results}
    \vspace*{-2mm}
\end{figure*}

\begin{figure}[!t]
	\vspace*{2mm}
	\centering
	\includegraphics[width = 0.995\linewidth]{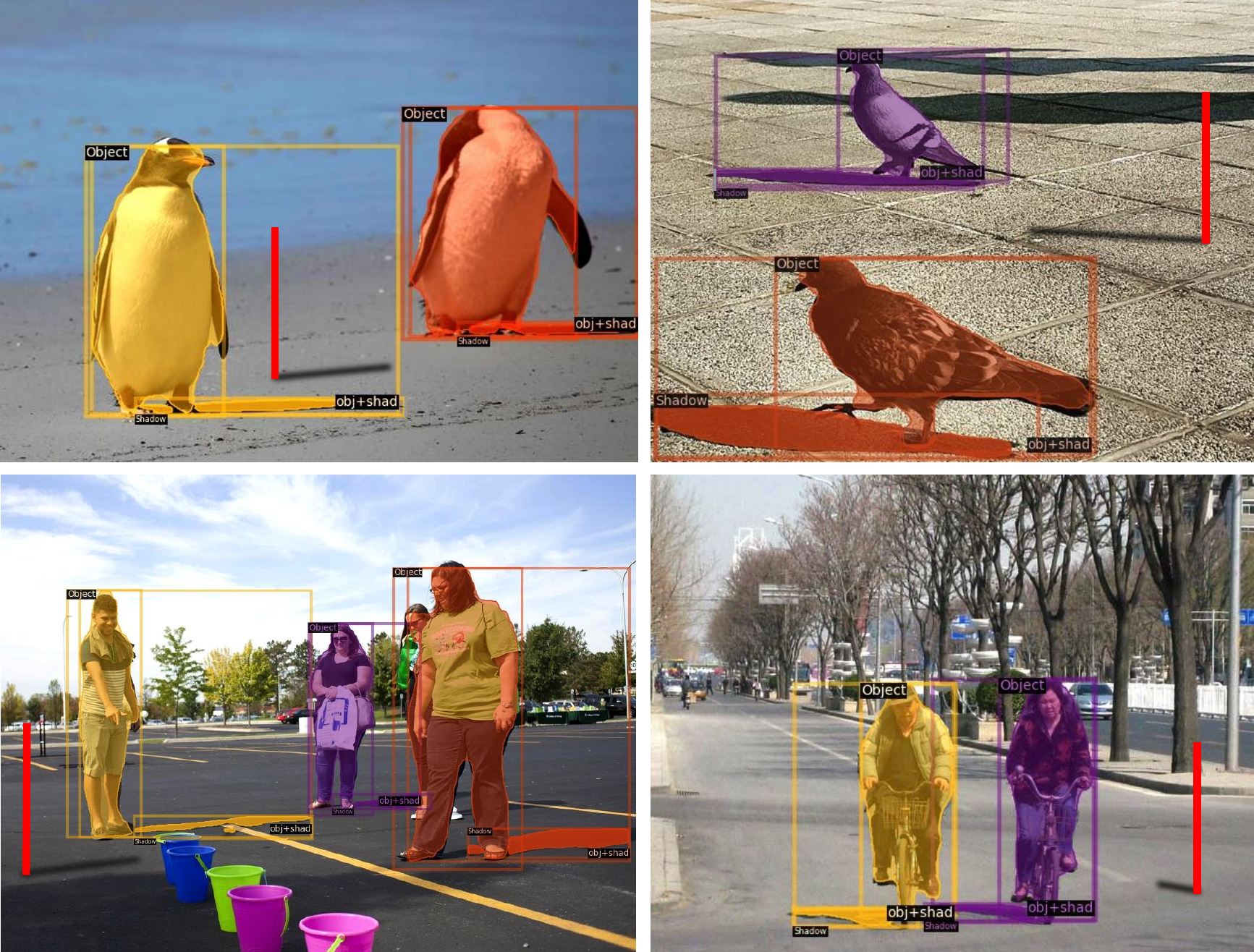} 
	\caption{Example images, where we estimate the light directions and incorporate virtual red posts with simulated shadows.}
	\label{img:application1}
    \vspace*{-3mm}
\end{figure}

%
%


\subsection{Results}




\paragraph{Evaluation.}
To evaluate the LISA framework, we set up 
(i) Baseline 1, which adopts only the top branch of LISA to predict bounding boxes and masks of the shadow and object instances, then merges them to form shadow-object associations based on the proximity between the shadow and object instances; and
(ii) Baseline 2, which removes the light direction head in LISA when predicting the shadow-object associations, but still adopts the procedure to pair-and-match the shadow and object instances (Section~\ref{sec:4.2}).


Tables~\ref{tab:SOAP_box} and~\ref{tab:SOAP_mask} report the quantitative comparison results in terms of the bounding boxes and masks in the final detected shadow-object associations.
Comparing different rows in the results, we can see that Baseline 2 clearly improves over Baseline 1, demonstrating that we can obtain better shadow-object associations in our deep end-to-end framework by independently predicting also the shadow-object associations and then pairing the shadow and object instances and matching them with the predicted shadow-object associations.
Moreover, by further predicting the light direction and taking it as the guidance to jointly optimize the framework, our full pipeline LISA achieves the best performance for all the evaluation metrics.
%

Figure~\ref{img:res1} shows visual comparison results for Baseline 1, Baseline 2, and our full pipeline.
The first column shows the input images, whereas the second, third, and fourth columns show the results produced by the two baselines and our full pipeline.
By comparing Baseline 1 with Baseline 2, we can see that further learning to detect the shadow-object associations independently in the deep framework helps to discover more shadow-object pairs, 
as shown in the third and fourth rows in Figure~\ref{img:res1}.
Moreover, after taking the light direction as guidance (Baseline 2 vs. full pipeline), our method improves the performance in various challenging cases,~\eg,
when there is large but irrelevant shadow region nearby (see the first row), 
when there are multiple shadow instances connect with a single object instance (see the second row), 
when the centers of the shadow and object instances are far from each other (see the third row), and
when there are multiple shadow regions near a single object instance (see the last row). 
Please see Figure~\ref{img:more_results} and supplemental material for more instance shadow detection results produced by our method on various types of images and objects.

\section{Applications}

%

Below, we present application scenarios to demonstrate the applicability of the results produced by our method.


\vspace*{-3mm}
\paragraph{Light direction estimation.}
%

First, instance shadow detection helps to estimate the light direction in a single 2D image, and we connect the centers of the bounding boxes of the shadow and object instances in each shadow-object association pair as the estimated light direction.
%
%
%
Figure~\ref{img:application1} shows some example results, where for each photo, we estimate the light direction and render a virtual red post with a simulated shadow on the ground based on the estimated light direction.
From the results, we can see that the virtual shadows with the red posts look consistent with the real shadows cast by other objects, thus demonstrating the applicability of our detection results.
%

%
 


\begin{figure}[!t]
	\centering
		\includegraphics[width = 0.484\linewidth]{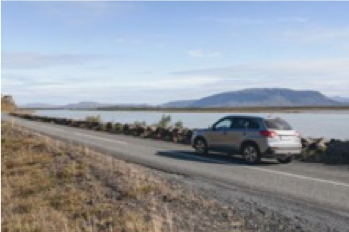}
		\includegraphics[width = 0.484\linewidth]{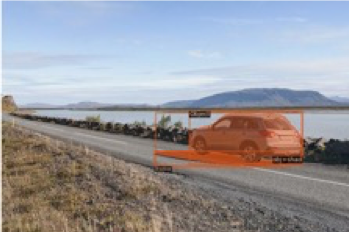}
	\begin{minipage}[t]{0.49 \linewidth}
		\vspace*{-3mm}
		\centerline{\footnotesize (a) Original image}
	\end{minipage}
	\begin{minipage}[t]{0.49 \linewidth}
		\vspace*{-3mm}
		\centerline{\footnotesize (b) Instance shadow detection}
	\end{minipage}

	\vspace{.8mm}
	\includegraphics[width = 0.484\linewidth]{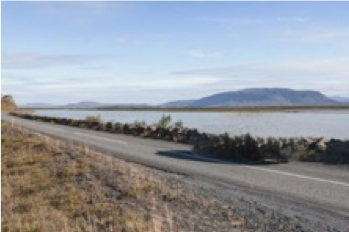}
	\includegraphics[width = 0.484\linewidth]{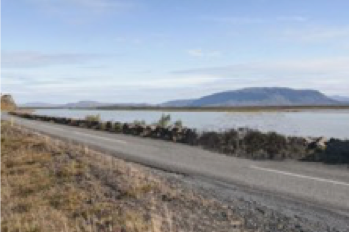}
	\begin{minipage}[t]{0.49 \linewidth}
		\vspace*{-3mm}
		\centerline{\footnotesize (c) An example result in~\cite{Uittenbogaard2019privacy}}
	\end{minipage}
	\begin{minipage}[t]{0.49 \linewidth}
		\vspace*{-3mm}
		\centerline{\footnotesize (d) Enhanced by our result}
	\end{minipage}

	\caption{Instance shadow detection enables us to easily remove objects (\eg, vehicle) with their associated shadows altogether.}
	\label{img:application2}
	\vspace*{-1.5mm}
\end{figure}

\vspace*{-3mm}
\paragraph{Photo editing.}
Another application to demonstrate instance shadow detection is photo editing, where we can remove not only the object instances but also their associated shadows altogether.
For privacy protection, Uittenbogaard~\etal~\cite{Uittenbogaard2019privacy} presents a method to automatically remove specific objects in street-view photos; see Figure~\ref{img:application2} (c) for a result, where it can successfully remove the vehicle.
However, the shadow cast by the vehicle remains on the ground.   
With the help of our instance shadow detection result (Figure~\ref{img:application2} (b)), we can remove the vehicle with its shadow altogether, as shown in Figure~\ref{img:application2} (d).

\begin{figure}[!t]
	\centering
		\includegraphics[width = 0.484\linewidth]{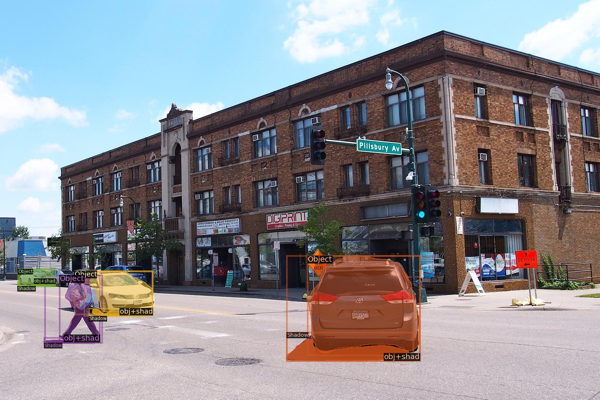}
		\includegraphics[width = 0.484\linewidth]{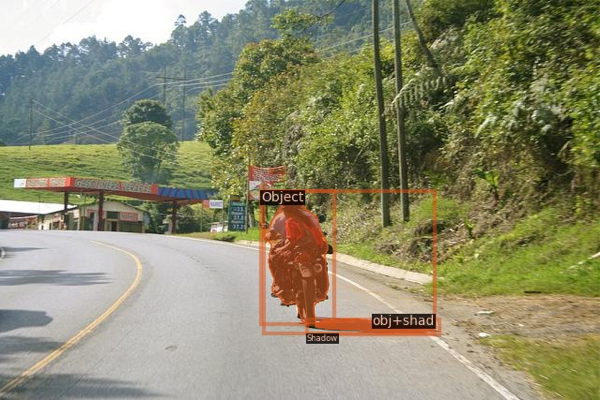}
	\begin{minipage}[t]{0.49 \linewidth}
		\vspace*{-3mm}
		\centerline{\footnotesize (a) Original image 1}
	\end{minipage}
	\begin{minipage}[t]{0.49 \linewidth}
		\vspace*{-3mm}
		\centerline{\footnotesize (b) Original image 2}
	\end{minipage}
	\vspace{0.8mm}
%
	
	\includegraphics[width = 0.484\linewidth]{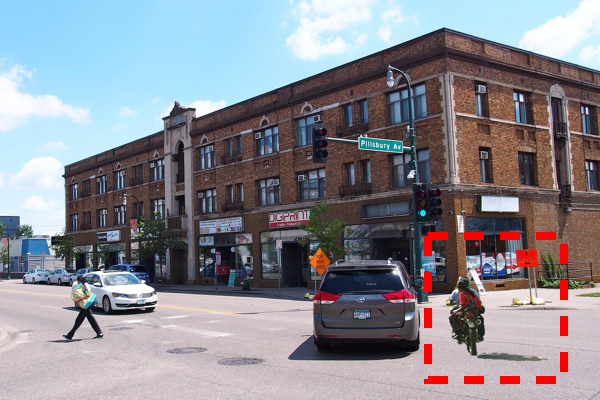}
	\includegraphics[width = 0.484\linewidth]{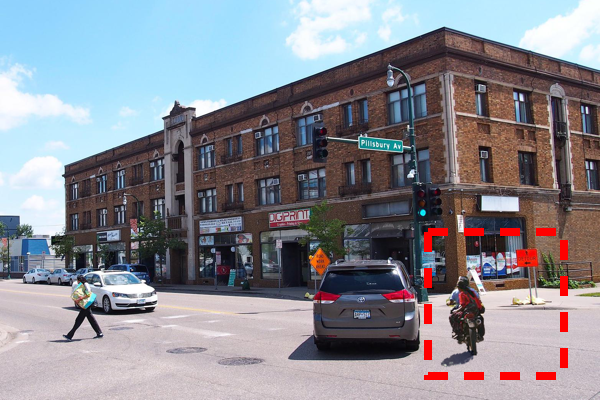}
	\begin{minipage}[t]{0.49 \linewidth}
		\vspace*{-3mm}
		\centerline{\footnotesize (c) Na\"ive cut-and-paste}
	\end{minipage}
	\begin{minipage}[t]{0.49 \linewidth}
		\vspace*{-3mm}
		\centerline{\footnotesize (d) Light-aware shadow}
	\end{minipage}
	\caption{When we cut-and-paste objects from one photo to the other, instance shadow detection results enable us not only to extract object and shadow instances together, but also to adjust the shadow shape based on the estimated light direction.}
	\label{img:application3}
	\vspace*{-1.5mm}
\end{figure}

Further, we can more efficiently transfer an object together with its shadow from one photo to another photo.
%
Figure~\ref{img:application3} presents an example, we cut the motorcycle with its shadow from (b) and paste them into (a) in smaller sizes.
Clearly, if we simply paste the motorcycle and shadow to (a), the shadow is not consistent with the real shadows in the target photo; see (c).
Thanks to instance shadow detection, which outputs individual masks for both object and shadow instances, as well as light directions.
Therefore, we can achieve light-aware photo editing by making use of the estimated light direction in both photos to adjust the shadow images when transferring the motorcycle from one photo to the other; see (d).



\section{Conclusions and Limitations}

In this paper, we presented instance shadow detection, which targets to find shadow instances and object instances, and pair them up together.
Also, we presented three technical contributions to approach the problem.
%
%
First, we prepare SOBA, a new dataset of 1,000 images and 3,623 pairs of shadow-object associations, where we provide the input photos together with a set of three instance masks.
Second, we develop LISA, an end-to-end deep framework, to predict boxes and masks of individual shadow and object instances, as well as boxes of shadow-object associations and the associated light directions; from these predictions, we further match the shadow and object instances, and pair them up to match with the predicted shadow-object associations and light directions for producing the output shadow-object pairs.
Third, we formulate SOAP, a new evaluation metric for quantitatively measuring the instance shadow detection results, enabling us to perform various experiments to compare with baseline frameworks.
In the end, we also demonstrate the applicability of our results on light direction estimation and photo editing.

As the first attempt to detect shadow-object instances, we admit that there are many possible methods that can be explored to improve the detection performance.
Besides methodologies, we did not consider the overlap between shadow instances associated with different objects.
Also, we did not consider cast shadows formed on some other object instances.
There are many open problems and unexplored situations for instance shadow detection.


In the future, we plan to first improve the performance of instance shadow detection by simultaneously leveraging multiple training data from the current datasets prepared for shadow detection and instance segmentation.
By exploring semi- or weakly-supervised methods to learn to detect instance shadows, we could combine the strengths and knowledge from various data to better the performance of instance shadow detection.
Last, we will also explore more applications based on the shadow-object association results.

\vspace*{-3mm}
\paragraph{Acknowledgments.} This work was supported by 
the Research Grants Council of the Hong Kong Special Administrative Region (Project no. CUHK 14203416),
the CUHK Research Committee Direct Grant for Research 2018/19, and
the Science and Technology Key Project of Guangdong Province, China (2019B010149002).


%

{\small
\bibliographystyle{ieee_fullname}
\bibliography{reference}
}

\end{document}